%% file: main.tex
\documentclass{article}

    \usepackage[preprint, nonatbib]{neurips_2023}

\usepackage{enumitem}

\usepackage{xcolor}
\definecolor{mint}{HTML}{00FFFF}
\definecolor{magenta}{HTML}{ff00ff}
\definecolor{green}{HTML}{00ff00}
\usepackage{newunicodechar}
\newunicodechar{−}{-}

\usepackage[utf8]{inputenc} 
\usepackage[T1]{fontenc}    
\usepackage{hyperref}       
\usepackage{url}            
\usepackage{booktabs}       
\usepackage{amsfonts}       
\usepackage{amsmath}        
\usepackage{nicefrac}       
\usepackage{microtype}      
\usepackage{xcolor}         
\usepackage{tcolorbox} 
\usepackage{tabularx}
\usepackage{booktabs}
\usepackage{graphicx} 
\usepackage{subcaption} 
\usepackage{cleveref}
\usepackage{natbib}
\usepackage{tikz}
\usepackage{amsmath}
\usepackage{amsfonts}

\usetikzlibrary{positioning}

\title{VRAIL: Vectorized Reward-based Attribution 
for Interpretable Learning}

\vspace{-2.5mm}
\author{
  Jina Kim\textsuperscript{1,*} \quad
  Youjin Jang\textsuperscript{1,*} \quad
  Jeongjin Han\textsuperscript{1, *} \\
  \textsuperscript{1}School of Computing, KAIST \\
  \texttt{\{jinakim, jyj22, hjj22\}@kaist.ac.kr}
}

\begin{document}
\maketitle

\begingroup
\renewcommand\thefootnote{*}
\footnotetext{Equal contribution}
\endgroup

\vspace{-2.5mm}
\input{contents/abstract}
\input{contents/main/introduction}
\input{contents/main/methods}

\input{contents/main/results}
\input{contents/main/analysis}

\input{contents/main/conclusion}

\clearpage
\bibliography{contents/reference}  

\appendix
\input{contents/appendix/contribution}
\input{contents/appendix/settings}
\input{contents/appendix/TNE}
\end{document}

%% file: contents/abstract.tex
\begin{abstract}
We propose \textbf{VRAIL} (Vectorized Reward-based Attribution for Interpretable Learning), a bi-level framework for value-based reinforcement learning (RL) that learns interpretable weight representations from state features. VRAIL consists of two stages: a deep learning (DL) stage that fits an estimated value function using state features, and an RL stage that uses this to shape learning via potential-based reward transformations. The estimator is modeled in either linear or quadratic form, allowing attribution of importance to individual features and their interactions. Empirical results on the Taxi-v3 environment demonstrate that VRAIL improves training stability and convergence compared to standard DQN, without requiring environment modifications. Further analysis shows that VRAIL uncovers semantically meaningful subgoals, such as passenger possession, highlighting its ability to produce human-interpretable behavior. Our findings suggest that VRAIL serves as a general, model-agnostic framework for reward shaping that enhances both learning and interpretability.

\end{abstract}

%% file: contents/main/introduction.tex
\vspace{-1.5em}
\section{Introduction}

In reinforcement learning (RL), designing an effective reward function remains a critical yet inherently challenging task. While the principle that “reward is enough” suggests that reward maximization alone can explain intelligent behavior \cite{rewardenough}, real-world environments often involve sparse, delayed, or ambiguous reward signals due to their complex and nuanced dynamics \cite{rl_challenge}. Humans make decisions by subjectively evaluating multiple factors, weighing changes in possession or outcomes against internal value judgments. For instance, consider the act of donating money: although it results in a loss of monetary possession, it may simultaneously produce a positive emotional outcome. Motivated by this insight, we introduce VRAIL (Vectorized Reward-based Attribution for Interpretable Learning), a novel method that approximates reward shaping through learnable feature weighting. VRAIL aims to accelerate agent convergence while making interpretability of learned values, by modeling how state transitions affect both possession and perceived value.

\begin{figure}[ht]
\vspace{-1em}
\centering
\begin{subfigure}[b]{0.45\linewidth}
    \centering
    \includegraphics[width=\linewidth]{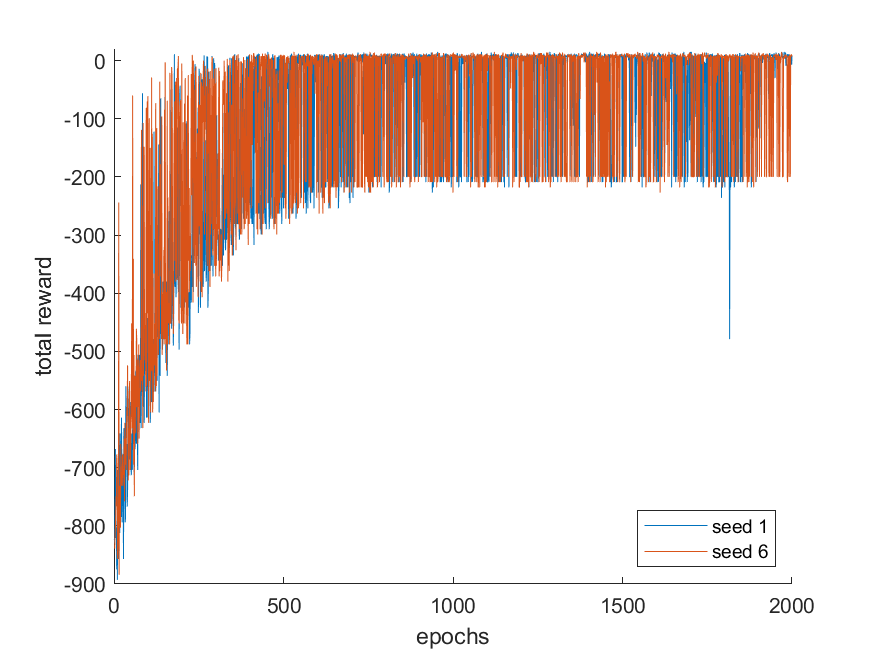}
    \caption{\textbf{DQN (w/o wall info)}}
    \label{fig:dqn_without_wall}
\end{subfigure}
\hfill
\begin{subfigure}[b]{0.45\linewidth}
    \centering
    \includegraphics[width=\linewidth]{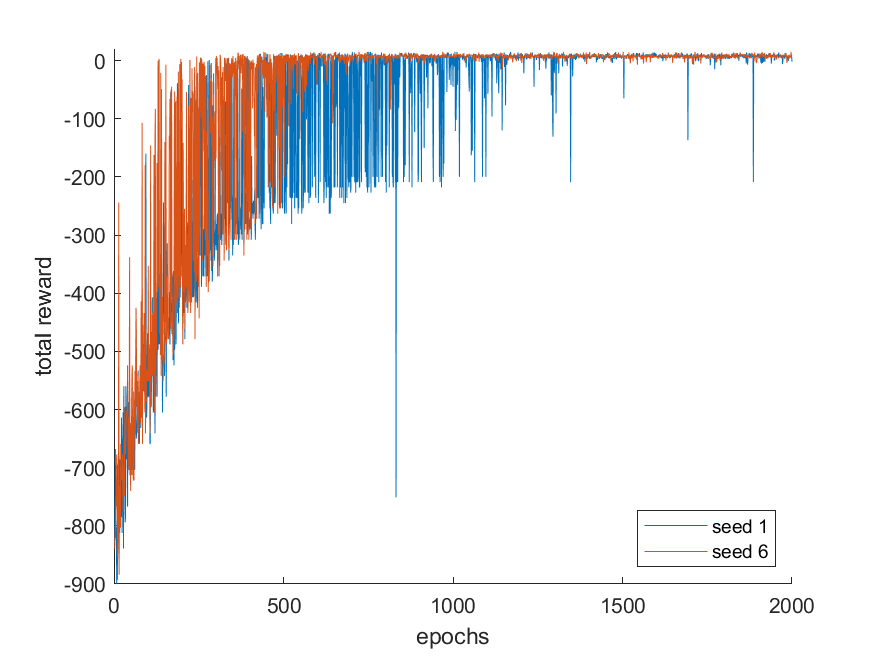}
    \caption{\textbf{DQN (w/ wall info)}}
    \label{fig:dqn_with_wall}
\end{subfigure}
\vspace{-0.5em}
\caption{\small \textbf{Effect of wall information on DQN convergence.} (a) Without wall information, DQN fails to converge in 2 out of 10 seeds. (b) Including wall information leads to consistent convergence across all seeds. For clarity, only the 2 failure cases are shown.}

\label{fig:taxi_dqn_wall_toy_exp}
\end{figure}

A toy experiment shown in~\cref{fig:taxi_dqn_wall_toy_exp} further illustrates the necessity of our proposed method. In the standard Taxi-v3 environment from the OpenAI Gymnasium library~\cite{gymnasium}, the agent’s observations do not include explicit wall information, which plays a critical role in successful navigation. As shown in \cref{fig:taxi_dqn_wall_toy_exp}, without this structural information, a vanilla DQN baseline fails to converge in 2 out of 10 random seeds, highlighting the instability that can arise from incomplete state representations. Although we found that manually adding wall information resolved the issue and led to consistent convergence, such interventions are impractical in real-world settings. Our expectation is that VRAIL achieves stable performance even in the original, unmodified environment by learning to attribute value to latent features within state transitions. Therefore, VRAIL will offer a practical and potentially generalizable approach for RL tasks where full state observability cannot be guaranteed.

%% file: contents/main/methods.tex
\section{Method}
\begin{figure}[ht]
  \centering
  \input{contents/figure/diagram}
  \caption{\textbf{Bi-level optimization framework of VRAIL.}  The DL stage learns reward parameters from state-value estimates, and the RL stage uses them to shape rewards and train the agent’s policy.}
  \label{fig:vrail_method}
\end{figure}
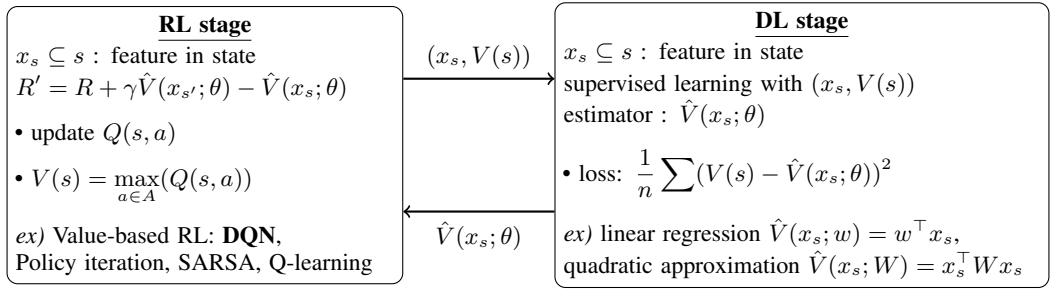

We propose \textbf{VRAIL} (Vectorized Reward-based Attribution for Interpretable Learning), a framework for learning interpretable weight via bi-level optimization. As illustrated in~\cref{fig:vrail_method}, the process consists of two interacting stages: a reinforcement learning (RL) stage and a deep learning (DL) stage. The overall objective is to help the agent recognize meaningful feature transitions as implicit subgoals, improving sample efficiency and interpretability.


\subsection{RL Stage: Reward Shaping with Learned Potentials}
\label{method:rl_stage}
In the RL stage, we train a value-based reinforcement learning agent using rewards shaped by the output of the DL stage. By the potential-based reward shaping theorem~\cite{reward_shaping_thm}, we can guarantee policy invariance under transformations of the form \[R'(s, a, s') = R(s, a, s') + \gamma \Phi(s') - \Phi(s).\]
For the shaping potential \( \Phi(s) \), we use an estimated value function $\hat V(x_s; \theta)$,  which approximates the expected cumulative return from \( x_s \subseteq s \), where $x_s$ is the extracted feature state vector \( s \) and $\theta$ represents the model parameter of the estimator.  
This choice is motivated by the observation that \( V(s) \) inherently encodes how close the agent is to achieving the main goal,  
thus implicitly capturing useful subgoal structures.  
By shaping the reward using \( V(s) \), we aim to encourage behaviors that guide the agent through intermediate progress toward the final objective.

\subsection{DL Stage: Learning a Interpretable Value Function}

In the DL stage, we make an estimator \( \hat {V}(x_s; \theta) \) to approximate the state-value function \( V(s) \). The function \( \hat {V}(x_s; \theta) \) is parameterized using either a linear form (Linear VRAIL) or a quadratic form (Quadratic VRAIL) as
\[
\hat {V}(x_s; w) = w^\top x_s \quad \text{or} \quad \hat {V}(x_s; W) = x_s^\top W x_s,
\]
where \( w \in \mathbb{R}^d \) and \( W \in \mathbb{R}^{d \times d} \) are learnable parameters.


\subsection{Bi-level Optimization Loop}


After each RL stage, the updated value function \( V(s) \) is collected and used to refine the reward parameters in the DL stage. This iterative procedure forms a closed bi-level optimization loop, where interpretable reward functions are continuously improved to support more effective learning.

%% file: contents/figure/diagram.tex
\begin{center}
\begin{tikzpicture}[node distance=1.3cm and 2.8cm, every node/.style={font=\small}]

\node[draw, rectangle, align=left, text width=5cm, rounded corners] (rl) {
    \centering\textbf{\underline{RL stage}}\\[0.2em]
    \raggedright
    $x_s \subseteq s$ : feature in state\\[0.2em]
    $R' = R + \gamma \hat V(x_{s'};\theta) - \hat V(x_s;\theta)$\\[0.8em]
    \textbullet\ update $Q(s, a)$\\[0.8em]
    \textbullet\ $V(s) = \max\limits_{a \in A}(Q(s,a))$\\[0.8em]
    \textit{ex)} Value-based RL: \textbf{DQN},\\
    Policy iteration, SARSA, Q-learning
};

\node[draw, rectangle, align=left, right=2cm of rl, text width=6.3cm, rounded corners] (dl) {
    \centering\textbf{\underline{DL stage}}\\[0.2em]
    \raggedright
    $x_s \subseteq s$ : feature in state\\[0.2em]
    supervised learning with $(x_s, V(s))$\\
    estimator : $\hat V(x_s; \theta)$\\[0.8em]
    \textbullet\ loss: $\displaystyle \frac{1}{n} \sum (V(s) - \hat V(x_s; \theta))^2$\\[0.8em]
    \textit{ex)} linear regression $\hat {V}(x_s; w) = w^\top x_s$,\\
    quadratic approximation $\hat {V}(x_s; W) = x_s^\top W x_s$
};

\draw[->, thick] ([yshift=25pt]rl.east) -- ([yshift=25pt]dl.west) node[midway, above] {$(x_s, V(s))$};

\draw[->, thick] ([yshift=-25pt]dl.west) -- ([yshift=-25pt]rl.east) node[midway, below] {$\hat V(x_s; \theta)$};

\end{tikzpicture}
\end{center}

%% file: contents/main/results.tex
\section{Experiments}
\subsection{Experimental Setup}
We conduct experiments on the classic Taxi-v3 environment~\cite{gymnasium}, where the agent navigates a gridworld to pick up and drop off a passenger. The environment provides discrete observations of the taxi’s position, passenger location, and destination. To evaluate VRAIL without modifying the original state space, we use the default environment without injecting features like wall information. While VRAIL supports various value-based RL algorithms~\cite{mnih2015dqn}, we focus on DQN for empirical evaluation. All models are run with 10 random seeds to assess convergence stability and robustness.

\subsection{Training Stability and Convergence Speed}
\begin{figure}[t]
\vspace{-1em}
\centering
\begin{subfigure}[b]{0.32\linewidth}
    \centering
    \includegraphics[width=\linewidth]{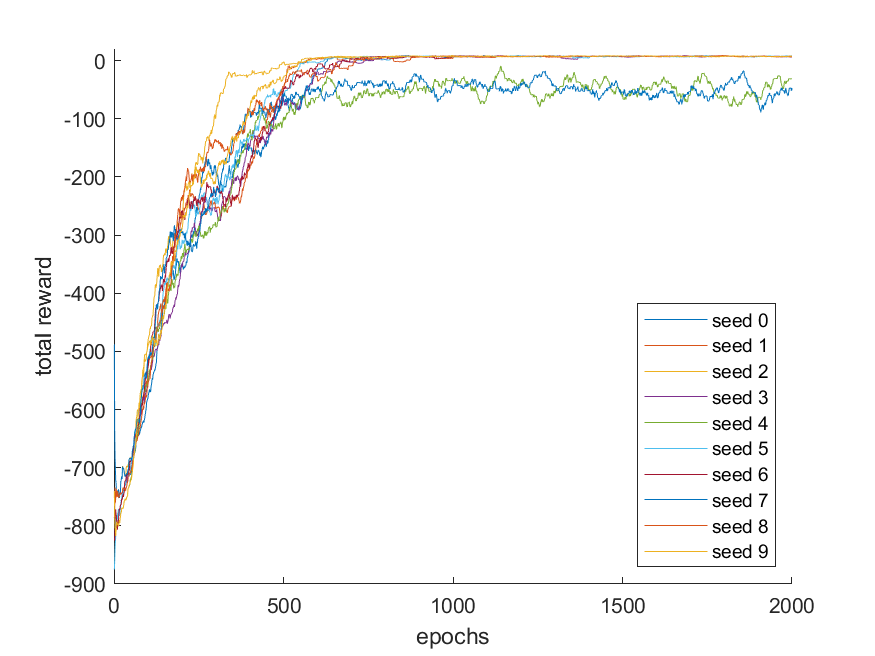}
    \caption{\textbf{DQN}}
    \label{fig:training_stability_dqn}
\end{subfigure}
\hfill
\begin{subfigure}[b]{0.32\linewidth}
    \centering
    \includegraphics[width=\linewidth]{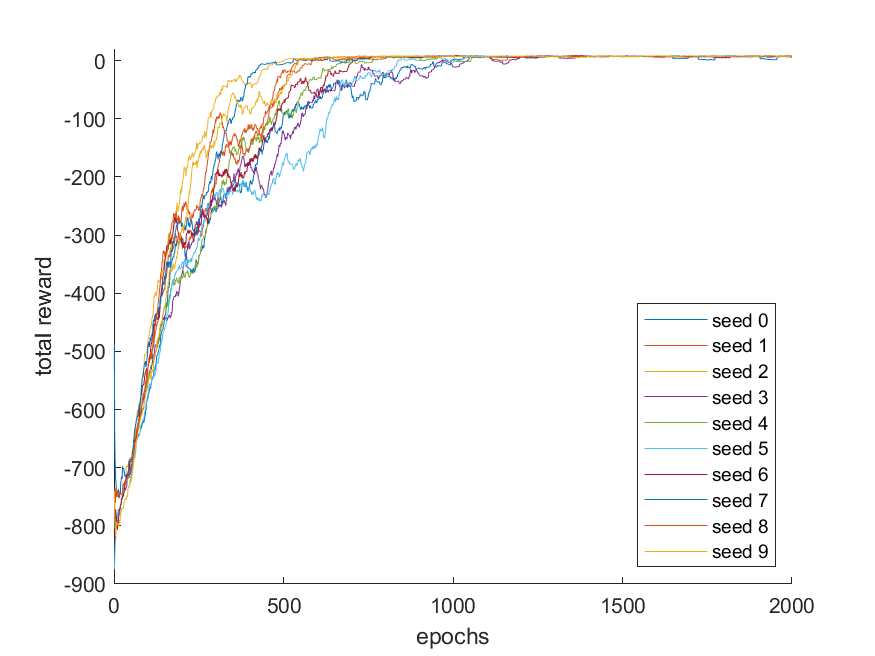}
    \caption{\textbf{Linear VRAIL (Ours)}}
    \label{fig:training_stability_linear}
\end{subfigure}
\hfill
\begin{subfigure}[b]{0.32\linewidth}
    \centering
    \includegraphics[width=\linewidth]{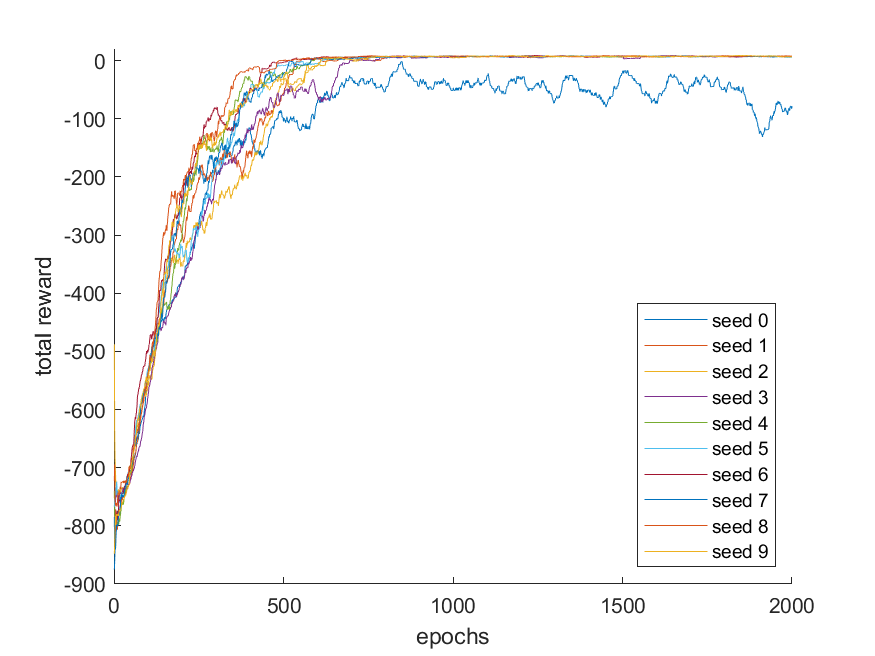}
    \caption{\textbf{Quadratic VRAIL (Ours)}}
    \label{fig:training_stability_quad}
\end{subfigure}
\vspace{-0.5em}
\caption{\small \textbf{Training stability comparison across models.} Total reward per episode over 2,000 training epochs. VRAIL variants (b, c) show improved convergence stability over the DQN baseline.}

\label{fig:training_stability}
\end{figure}
We evaluate the convergence behavior of VRAIL in terms of both training stability and speed, using a standard DQN baseline for comparison. A run is considered converged if the agent consistently achieves a moving average reward above a fixed threshold during training. As shown in~\cref{fig:training_stability}, the baseline DQN converged in 8 out of 10 runs. In contrast, Linear VRAIL converged in all 10 runs and  Quadratic VRAIL converged in 9 out of 10 runs, demonstrating greater robustness than the baseline.

\begin{table}[h]
\caption{Average epochs to reach reward thresholds (10 seeds, excluding top/bottom 2 outliers).}
\vspace{0.5em}
\hspace*{0.1cm} 
\begin{minipage}{\dimexpr\linewidth-1cm} 
\centering
\footnotesize

\begin{tabular}{lcccc}
\toprule

\textbf{Reward Threshold} & $-10$ & $-5$ & $0$ & $+5$ \\
\midrule
DQN & 600.00 & 612.17 & 648.17 & 717.67 \\
Linear VRAIL (Ours) & 614.17 & 643.17 & 652.50 & 735.83 \\
\textbf{Quadratic VRAIL (Ours)} & \textbf{538.17} & \textbf{562.83} & \textbf{594.33} & \textbf{660.17} \\
\bottomrule
\end{tabular}
\label{tab:reward_threshold_epochs}
\end{minipage}
\end{table}
We evaluate convergence speed by measuring the number of training epochs required to reach moving average reward thresholds of $-10$, $-5$, $0$, and $+5$. As shown in~\cref{tab:reward_threshold_epochs}, Quadratic VRAIL consistently reaches all thresholds faster than both DQN and Linear VRAIL. While Linear VRAIL converges more reliably, it takes slightly longer than DQN to achieve the same rewards. These results highlight a trade-off: Linear VRAIL offers robustness, while Quadratic VRAIL combines speed with stability.

\subsection{Effect of Shaped Reward Transfer}

\begin{figure}[ht]
\vspace{-1em}
\centering
\begin{subfigure}[b]{0.45\linewidth}
    \centering
    \includegraphics[width=\linewidth]{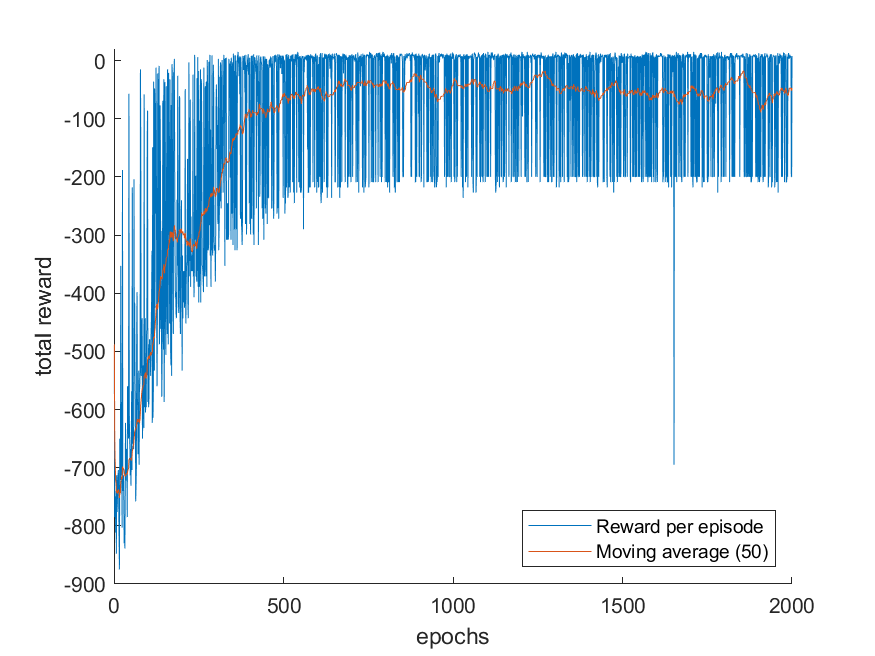}
    \caption{\textbf{DQN}}
    \label{fig:dqn_default}
\end{subfigure}
\hfill
\begin{subfigure}[b]{0.45\linewidth}
    \centering
    \includegraphics[width=\linewidth]{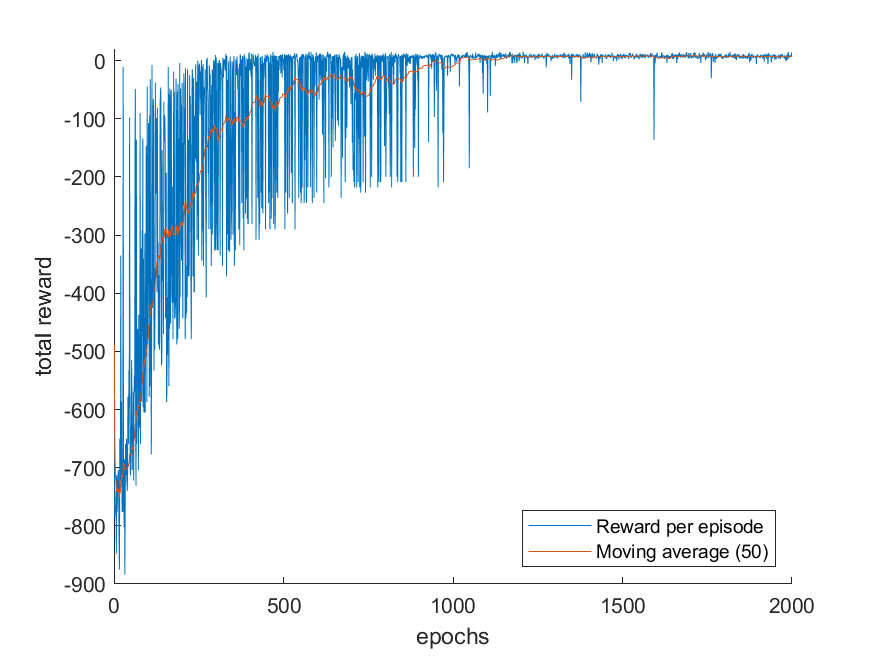}
    \caption{\textbf{DQN (w/ trained $w$)}}
    \label{fig:dqn_vrail_applied}
\end{subfigure}
\vspace{-0.5em}
\caption{\small \textbf{Effect of reward shaping using a pretrained Linear VRAIL model on DQN training.} Standard DQN fails to converge in 1 out of 5 seeds, while DQN trained with the reward function learned by Linear VRAIL converges in all 5 runs.}

\label{fig:dqn_vs_dqn_vrail_applied}
\end{figure}
To further isolate the effect of learned reward shaping, we evaluate whether the reward function learned by VRAIL can benefit standard RL agents when applied independently. Specifically, we pretrain Linear VRAIL to learn a potential-based reward function, then fix this function and apply it to a vanilla DQN agent without further reward updates.
As shown in~\cref{fig:dqn_vs_dqn_vrail_applied}, the baseline DQN fails to converge in 1 out of 5 seeds. In contrast, DQN trained with the reward shaped by pretrained Linear VRAIL successfully converges in all 5 seeds. This suggests that VRAIL’s learned reward function serves as a robust shaping mechanism, improving convergence stability even when decoupled from the full bi-level optimization loop.


%% file: contents/main/analysis.tex
\section{Analysis}
To highlight the interpretability of VRAIL, we analyze the learned parameters in our models.

\begin{figure}[ht]
\centering
\begin{subfigure}[b]{0.45\linewidth}
    \centering
    \includegraphics[width=\linewidth]{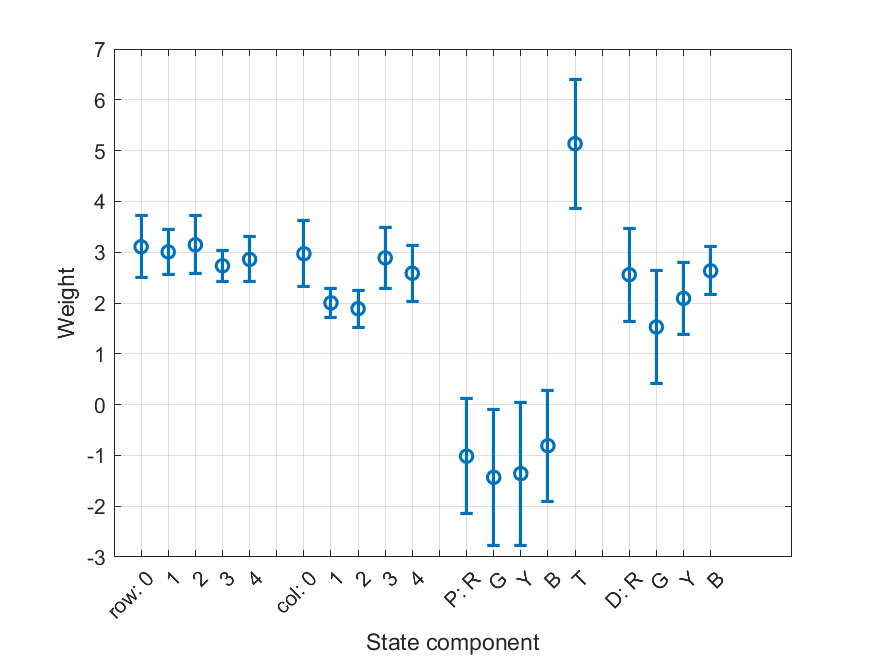}
    \caption{\textbf{Linear VRAIL ($w$)}}
    \label{fig:linear_weight}
\end{subfigure}
\hfill
\begin{subfigure}[b]{0.45\linewidth}
    \centering
    \includegraphics[width=\linewidth]{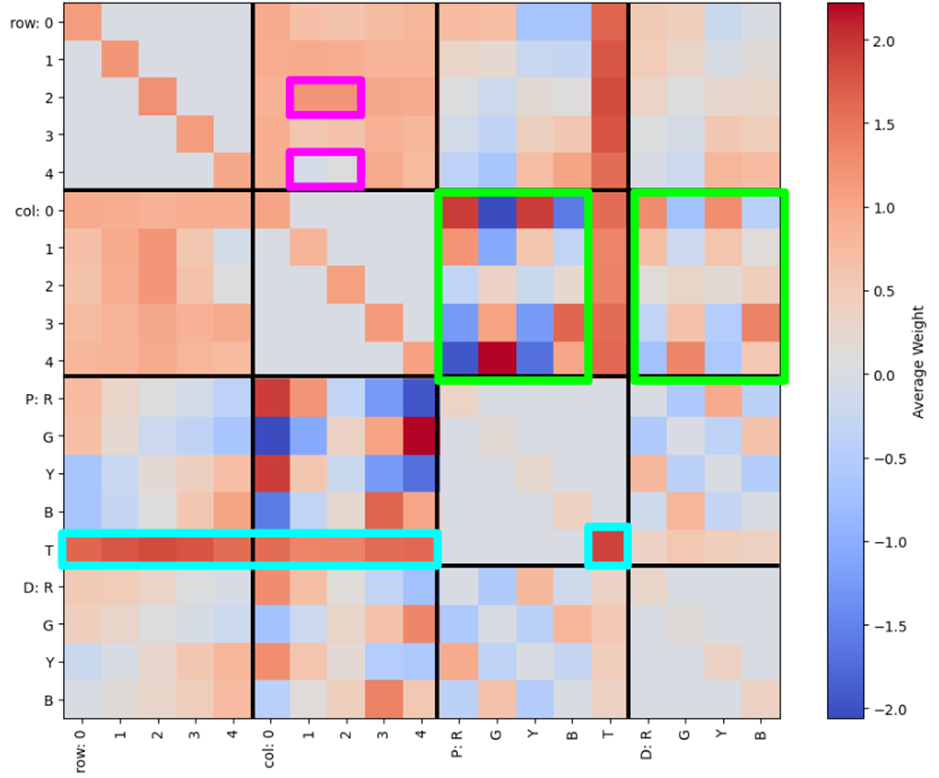}
    \caption{\textbf{Quadratic VRAIL ($W$)}}
    \label{fig:quadratic_weight}
\end{subfigure}
\vspace{-0.5em}
\caption{\small \textbf{Visualization of learned parameters of DL stage.} (a) Linear VRAIL contains the importance of each feature, (b) while Quadratic VRAIL contain the pairwise feature importance.}
\end{figure}

\subsection{Linear VRAIL}

In Linear VRAIL, the value function is represented as $\hat {V}(x_s; w) = w^\top x_s$, where \( w \in \mathbb{R}^{19} \) is a weight vector, with each element corresponding to a state feature. To capture a range of learned behaviors across different runs, we average the learned weight vectors over 10 random seeds and visualize the resulting feature importance. As shown in~\cref{fig:linear_weight}, the feature assigned the highest weight corresponds to whether the passenger is currently in the taxi, represented as "P: T". This suggests that the model implicitly identifies \textit{passenger possession} as a key subgoal. Notably, this aligns with the structure of the Taxi environment, where incorrect pickup or drop-off actions incur a penalty of \(-10\), thereby reinforcing proper pickup and drop-off.

\subsection{Quadratic VRAIL}

In Quadratic VRAIL, the value function is defined as \(\hat{V}(x_s; W) =   x_s^\top W x_s \),  
where \( W \in \mathbb{R}^ {19 \times 19} \) captures pairwise interactions between state features.  
We averaged the learned $W$ over 9 converged seeds and visualized it as a heatmap. In~\cref{fig:quadratic_weight}, each axis represents a state feature, and cell intensity reflects the strength of interaction between feature pairs. 

In the \textcolor{magenta}{\textbf{magenta}} box, we observe interactions between spatial (row, column) features.  
Higher weights at accessible positions (e.g., (2,1)) indicate more frequent movement, while lower weights at blocked positions (e.g., (4,1)) reflect restricted mobility. In the \textcolor{green}{\textbf{green}} box, we observe strong interactions between column features and both passenger and destination locations.  
Although both show similar patterns, the passenger-related interactions are consistently stronger, suggesting that the agent prioritizes \textit{pickup} before heading to the destination. Finally, the \textcolor{mint}{\textbf{mint}} box highlights interactions involving the ``P: T'' feature (passenger is in the taxi).  
This region exhibits consistently high weights across multiple feature combinations,  indicating that \textit{passenger possession} plays a central role in shaping the agent’s policy.

%% file: contents/main/conclusion.tex
\section{Conclusion}

We introduced VRAIL, a vectorized reward attribution method aimed at enhancing both learning effectiveness and interpretability in reinforcement learning. Built on a bi-level optimization framework, VRAIL shapes rewards using feature-based value function estimators, enabling agents to identify meaningful subgoals without modifying the environment. Experiments in the Taxi-v3 domain demonstrate that VRAIL improves both convergence stability and learning speed. Analysis of the learned parameters further reveals interpretable patterns in agent behavior, such as prioritizing passenger possession. These results highlight VRAIL’s potential for real-world RL scenarios with sparse or delayed rewards. As future work, we plan to extend our evaluation to more complex and partially observable environments such as MicroRTS and Crafter, to further assess VRAIL’s generalizability and effectiveness in dynamic, high-dimensional settings.

%% file: contents/appendix/contribution.tex
\clearpage
\section{Team member Contribution}
All team members actively contributed to every stage of the project, including project concretization, trials for method improvement, slides, and report writing. Further individual contributions are detailed below.

\begin{itemize}[leftmargin=1em]
\item Jina Kim : Backbone model implementation (Linear VRAIL), DQN toy experiment (Fig 1), self-attention based model trials, final refinement of the report.

\item Youjin Jang : Idea proposal (vectorized rewards), presentation, quiz creation, visualization (graphs, diagrams).

\item Jeongjin Han : Interpretation of learned parameters of DL stage, effect of shaped reward experiment (Fig 4), $\alpha$-scheduling trials, environment variants trials. 
\end{itemize}

%% file: contents/appendix/settings.tex
\section{Hyperparameter Settings}
We set the number of outer optimization cycles to 20 and use a discount factor of $\gamma = 0.99$. The exploration rate $\epsilon$ decays as $\epsilon(t) = \max(0.995^t, 0.01)$. In the RL stage, we use DQN with experience replay. The agent is trained for 100 epochs with a learning rate of $1\mathrm{e}{-3}$, batch size 64, and replay buffer size 50,000. The target network is updated every 10 epochs. In the DL stage, we train for 50 epochs with a learning rate of $1\mathrm{e}{-2}$.




%% file: contents/appendix/TNE.tex
\section{Trial and Error}

We conducted a series of trial-and-error experiments to explore the effects of various design choices in reward shaping and value estimation in VRAIL. Our study considered the following four dimensions:

\begin{itemize}[leftmargin=1em]
    \item \textbf{Environment variants:}  Sparse reward version (Only +20 when arrival, $-10$ for illegal actions) was tested, but didn't work well at all.

    \item \textbf{State value estimators:}  We explored a self-attention-based model design, but due to the absence of an additional dimension over which to apply softmax, we switched to a quadratic model.

    \item \textbf{Reward shaping bonuses:}
    \begin{itemize}
        \item Global shaping using $\alpha\times\tanh(w^\top x)$
        \item Element-wise shaping: $\alpha\times\sum_i \tanh(w_i x_i)$
    \end{itemize}
    These methods worked well for some seeds, but showed instability across others, likely because this form of shaping alters the agent’s ultimate objective.

    \item \textbf{Alpha scheduling:} Constant, linear, sigmoid, exponential, and logarithmic schedules were tested to control the mixing coefficient $\alpha$. These schedules showed meaningful improvements, but our latest model no longer requires \( \alpha \).
\end{itemize}